\pdfoutput=1

\documentclass[11pt]{article}

\usepackage[]{EACL2023}

\usepackage{times}
\usepackage{latexsym}

\usepackage[T1]{fontenc}

\usepackage[utf8]{inputenc}

\usepackage{microtype}

\usepackage{inconsolata}

\usepackage{adjustbox}
\usepackage{multicol}
\usepackage{multirow}

\usepackage{graphicx}
\usepackage{microtype}
\usepackage{footmisc}
\usepackage{ctable}

\usepackage[labelfont=bf]{caption}
\usepackage{subcaption}
\usepackage{mathrsfs}
\usepackage{bbding}
\usepackage{pifont}
\usepackage{soul}

\newcommand{\vabart}{{\scshape VA-BART}}
\newcommand{\varoberta}{{\scshape VA-RoBERTa}}
\newcommand{\vaalbert}{{\scshape VA-ALBERT}}

\newcommand{\ftbart}{{HL-BART}}
\newcommand{\ftroberta}{{HL-RoBERTa}}
\newcommand{\ftalbert}{{HL-ALBERT}}
\newcommand{\ftmodels}{{HL-Models}}

\newcommand{\ours}{{\scshape VA-model}}
\newcommand{\specificthanks}[1]{\@fnsymbol{#1}}

\title{Enabling Classifiers to Make Judgements\\Explicitly Aligned with Human Values}

\author{
Yejin Bang$^1$\thanks{\quad~Equal contribution.}\quad
Tiezheng Yu$^1$$^*$\quad
Andrea Madotto$^2$\\
\bf{Zhaojiang Lin$^2$\quad
Mona Diab$^2$\quad
Pascale Fung$^1$\thanks{\quad~The author contributed to the original idea as a part of responsible AI project for Meta AI.}} \\
$^1$The Hong Kong University of Science and Technology  \quad $^2$Meta AI \\
\texttt{\{yjbang,tyuah\}@connect.ust.hk}
}

\begin{document}
\maketitle
\begin{abstract}
Many NLP classification tasks, such as sexism/racism detection or toxicity detection, are based on human values. Yet, human values can vary under diverse cultural conditions. Therefore, we introduce a framework for value-aligned classification that performs prediction based on explicitly written human values in the command. 
Along with the task, we propose a practical approach that distills value-aligned knowledge from large-scale language models (LLMs) to construct value-aligned classifiers in two steps.
First, we generate value-aligned training data from LLMs by prompt-based few-shot learning. Next, we fine-tune smaller classification models with the generated data for the task. Empirical results show that our \ours{s}~surpass multiple baselines by at least $15.56$\% on the F1-score, including few-shot learning with OPT-175B and existing text augmentation methods. We suggest that using classifiers with explicit human value input improves both inclusivity \& explainability in AI. 
\end{abstract}

\section{Introduction}
\label{sec:Introduction}
\begin{figure}[t]
    \centering
    \includegraphics[width=\linewidth]{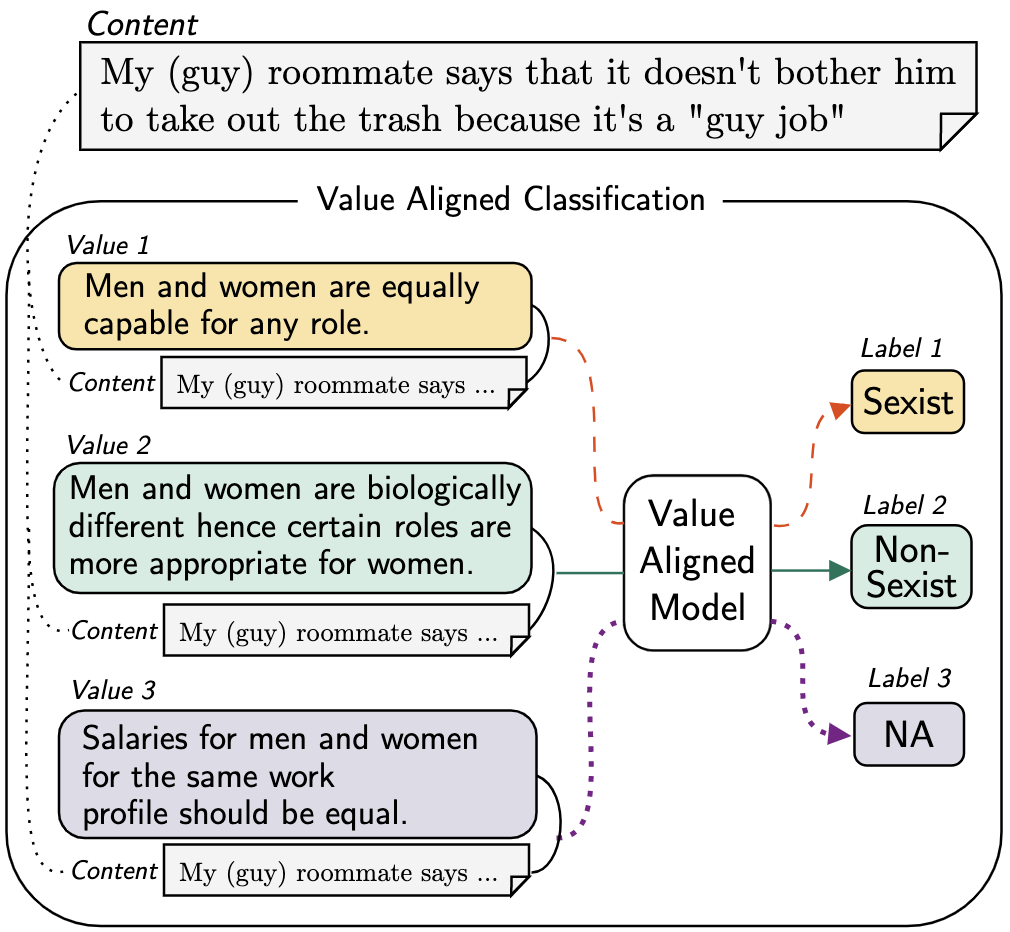}
    \caption{Illustration of proposed value alignment task. Given the same content, \ours~makes variable predictions based on explicitly provided human values. 
    }
    \label{fig:introduction}
\end{figure}

The demand for responsible NLP technology -- to make it more robust, inclusive and fair, as well as more explainable and trustworthy -- has increased since pre-trained large-scale language models (LLMs) have brought about significant progress in making NLP tasks more efficient and broad-ranging \cite{brown2020language,zhang2022opt,chowdhery2022palm, radford2019language,brown2020language,petroni2019language,madotto2020language}. Researchers have studied how to align machines with human values as one of the directions to achieve responsible AI technology by teaching machines about moral and social norms \cite{forbes2020social, emelin2020moral, jiang2021delphi}, ethics and common human values \cite{hendrycks2020aligning} or human preferences \cite{christiano2017deep, koren2008factorization}.

Value-alignment of AI systems is not a trivial problem as human values are non-consensual by nature~\cite{hanel2018cross}. 
Values can be very diverse and most existing works have attempted to align machines with shared human values or average norms, or from a certain cultural perspective with crowd sourced annotations \cite{jiang2021delphi}.
These days, for instance, many societies agree that sexism should be eliminated, and we expect machines to be non-sexist, but different individuals and cultures may perceive sexism differently. As is shown in Figure \ref{fig:introduction}, the same content can be considered to be sexist or non-sexist depending on the values provided to make the judgements.



In this paper, we propose a value-aligned judgement task that separates the value definition process from the development of the models for more inclusive and explainable value-aligned NLP. Our proposed task aims to build a single model to make dynamic judgements based on explicitly provided human values, requiring the model to understand the value and its corresponding entailment on the given content. The value is provided in the form of instructions, allowing coarse-to-fine customization.  We start with value-aligned sexism classification as a proof of concept for the proposed approach, as sexism is one of the most representative examples of varying cultural perspectives.


We also present Value-Aligned Models (\ours{s}) that leverage value-based knowledge from LLMs.
LLMs are trained from vast amounts of human data with embedded human values \cite{hendrycks2020aligning}. However, they are not controllable and it is difficult to fine-tune such large models with explicit value alignment. 
Instead,  we distill value-based training data from the LLMs using prompt-based data generation with example values, and build \ours~ by fine-tuning smaller classification models with the distilled data. 
Experimental results show that our approach is more stable and accurate than directly applying few-shot learning on LLMs. Moreover, our methodology avoids costly human labeling or crowd sourcing of values, allowing easier extensions to other value-aligned tasks in different domains.
We further investigate model performance using data generated from different scales and types of LLMs, and study the effect of data size for fine-tuning, and analyze the quality of the generated data. Moreover, we study the generalization ability of \ours{s} by testing its performance on unseen value sets.

Our contributions are as follows: 1) we introduce the value-aligned classification task, where we first define human values externally and then use them at the instruction level in an in-context learning paradigm and construct value-aligned classifiers to make predictions; 2) we propose to leverage prompt-based data generation to distill value-aligned knowledge from LLMs for smaller classification models; 3) experimental results indicate that our approach significantly outperforms strong baselines, including in-context few-shot learning with LLMs and existing text augmentation methods; 4) we systematically study factors that impact prompt-based data generation and highlight research questions and challenges in the value-aligned judgement task through thorough analysis. 



\section{Related Work}
\paragraph{Human Value Alignment}
One challenge in value alignment is value definition, and there has been a profusion of documents on AI ethical standards \cite{gabriel2020artificial, dignum2017responsible}. \citet{jobin2019global} identified 11 clusters of ethical principles among 84 documents, and \citet{fjeld2020principled} found eight key themes across 36 of the most influential of them. However, since human values are variable with culture, we anticipate value definition to be dynamic. Meanwhile, the values should be defined externally to the development of the NLP algorithms, like how we adopt definitions of sexism categories based on social studies. 

To teach models value-alignment, the literature has focused on improving the model's reasoning ability relating to human values and morality \citep{forbes2020social,emelin2020moral,lourie2021scruples, hendrycks2020aligning}.
Recently, \citet{solaiman2021process} proposed to fine-tune GPT-3 to adapt to a manually crafted values-targeted dataset to arrive at a values-targeted model. However, in their approach, value alignment and definition are intertwined and entangled in an iterative process. We instead separate the value definition and alignment process models about value-aligned judgement with explicit value provision.


\paragraph{Prompt-based Learning}
Recently, LLMs have shown great performance on prompt-based learning \cite{brown2020language,chowdhery2022palm}, which doesn't require fine-tuning. Instead, the model is directly fed a prompt that includes some examples, and the model can generate results as if it has ``learned''. Studies on efficient prompt-learning/-construction include \citet{lu2021fantastically, reynolds2021prompt, zhao2021calibrate, schick2020few}. We consider the literature for prompt-construction in our methodology.

\paragraph{Knowledge Distillation} Knowledge distillation is the transfer of knowledge from teacher to student distribution \cite{distilling2015hinton}. Recent works have attempted to perform distillation from LLMs by prompting for text generation to show that it outperforms existing text augmentation methods \cite{yoo2021gpt3mix, wang2022exploring}.  \cite{west2021symbolic} retrieves commonsense knowledge symbolically in a text form from GPT-3 for downstream tasks with help of smaller filtering classifiers. We distill \textit{value-specific} knowledge, not all abilities of general language model, from LLMs through value-aligned training data generation for training smaller value-aligned classifiers. This reduces the cost of human labeling and also enables building smaller models specialized for value-aligned judgment task.




\section{Value-Aligned Judgement Task}
\label{sec:task_introduce}
\subsection{Task Description}
As an effort to align machines with human values, our task focuses on teaching the model that different values can lead to different judgements even given the same content. The task is formulated as follows. A model needs to make a judgement $Y_V$ on content $C$ based on an explicit human value $V$. In this work, ``value'' refers to any qualities, standards of behavior, or beliefs that individuals or societies hold, and is expressed in natural language phrases or sentences. The set of values is externally defined by a human user of the system or from existing relevant literature on moral philosophy, and is independent of the development of algorithms. The distinction from the existing value-aligned classification task
and conventional classification tasks is that our task expects the model to incorporate 
\textit{explicitly provided values} along with other inputs for making judgements. 


We separate the process of value definition from the development of the value-aligned models so that the models can learn to make dynamic judgements based on external values. For instance, existing sexism classifiers implicitly learn a fixed set of definitions of sexism from labeled data, so the content will be judged based on these static values. Our task requires the model to predict dynamic labels depending on the different explicit values even when the content is the same. 

\subsection{Value-aligned Sexism Classification}
We showcase the value-aligned judgement task with an application to sexism classification. The model needs to judge whether natural language content is sexist or non-sexist based on a given value $V$. If the value is not applicable or irrelevant, the model needs to predict that it is not applicable (NA). 
Our rationale for choosing the sexism classification task is that the definition of sexism has changed over time as values have evolved and altered and it still varies across cultures. Thus, we can verify the effect of varying values in a more evident manner in the sexism classification task. Furthermore, the importance of a fine-grained understanding of sexism has been emphasized~\cite{jha-mamidi-2017-compliment, sharifirad-etal-2018-boosting, parikh-etal-2019-multi}.
This aligns with our motivation for explicit value-aligned judgement. Lastly, values related to sexism are complicated, involving religious, cultural, and personal beliefs or values. We thus believe it is a task with enough complexity to act as a case study. 



\section{Methodology}
\label{sec:methodology}
There is no existing resource for training value-aligned classification models. We therefore propose to leverage LLMs for generating synthetic training data. LLMs have been found to learn significant amounts of inherent knowledge as well as human values during pre-training \cite{petroni2019language, hendrycks2020aligning, west2021symbolic, roberts2020much}. 
However, the direct usage of LLMs in zero-shot setting for NLP tasks can be unstable and still limited~\cite{wei2021finetuned}.
The richly embedded knowledge in LLMs nevertheless makes them good resource generators. Therefore, we attempt to build value-aligned models (\ours{s}) through fine-tuning smaller models on the value-aligned training data generated by LLM(s).

\begin{figure}[t]
    \centering
    \includegraphics[width=0.95\linewidth]{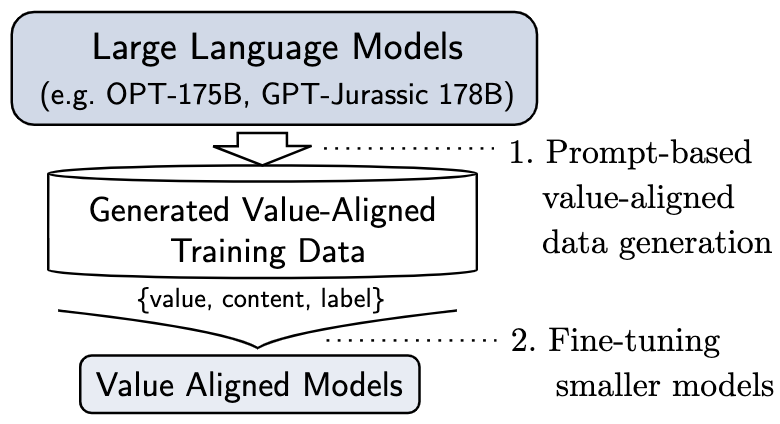}
    \caption{Illustration of the construction of our proposed  \ours. Using LLMs, we first create synthetic value-aligned training data. Then, we transfer the knowledge into smaller models by fine-tuning them on the data, so Value Aligned Models can make value-aligned judgements.}
    \label{fig:pipeline}
\end{figure}

Our proposed method (Figure \ref{fig:pipeline}) consists of two steps: 1) prompting human value-aligned contents from LLMs by providing explicit human values and instructions, and 2) fine-tuning smaller LMs on the generated data to teach them about value-aligned judgements. 
Formally, we build \ours~ (parameterized by $\theta$) to maximize the following likelihood:
\begin{equation}
    L(\theta) = logP(Y|V,C;\theta).
\end{equation}



\subsection{Value-Aligned Knowledge Distillation: Prompt-based Data Generation}
\label{sec:data_generation}

\paragraph{Prompt Construction with Few-shot Examples}
The prompt construction of in-context few-shot examples affect performance. Thus we refer to the existing literature on different prompt-techniques \cite{reynolds2021prompt,zhao2021calibrate, yoo2021gpt3mix}. For the few-shot examples, we create a pool of 10 human-labeled samples (value, content, and value-aligned labels) for each value. According to \citet{lu2021fantastically}, the order of the few-shot samples in the prompt affects the in-context learning for LLMs. Therefore, we randomly select and order five samples out of the pool.

To select the most appropriate prompt for generating value-aligned synthesized data, we test five candidate prompt templates with reference to literature. All prompt templates consist of a label, a value, and value-aligned content examples. The best-performing prompt template is selected based on testing with a smaller size of the samples. The prompt templates and their performance are available in Appendix~\ref{appendix:Prompt_Choice_and_Results}.


\paragraph{Generation}
We feed the prompts to LLMs to generate value-aligned synthetic training samples. Our method is model agnostic in that any LLMs can be adopted for this step. Recently, LLMs have scaled to more than 500 billion parameters \cite{chowdhery2022palm,smith2022using}, and some models with more than 100 billion parameters are available publicly, such as Jurassic-1 Jumbo (GPT-Jurassic~\citet{lieber2021jurassic}), Open Pre-trained Transformer (OPT~\citet{zhang2022opt}), and GPT-3. In this paper, we choose OPT-175B for the main experiment and provide an analysis on the effects of the size and types of LLMs.




\paragraph{Generated Content Extraction \& Processing}
The generated content is generated in succession after the prompt as natural text, and extracted through pattern matching. We gather all extracted content to construct a synthetic training set for teaching the smaller models in the next step, and process the generated data as follows. Firstly, we keep only unique samples by dropping all duplicates. Then, we remove exact copies of the few-shot examples used in the prompts. Finally, any content less than three words is filtered out as it is less informative. 

\subsection{Fine-tuning Smaller Models -- Value-Aligned Models}
\label{sec:Fine-tuning_Classifiers}
In the next phase, we build classifiers by fine-tuning relatively smaller transformer-based models (e.g., ALBERT-base, RoBERTa-base, BART-base) with the generated training data to enable them to make value-aligned judgements. We add a linear layer on top of the pooled output of the smaller models to construct our proposed \ours. In order to make the model intake both values and content in the learning phase, the input text is formatted into \texttt{``value [sep] content [sep]''} and the output is a value-aligned judgement.

The classifiers need to predict different labels according to explicitly provided values given the same content. Recalling the example of value-aligned sexism classification in Figure \ref{fig:introduction}, the same content can be considered to be sexist, non-sexist or NA depending on the considered values. 

\section{Experiments}
In this paper, we conduct value-aligned sexism classification. Models are expected to label content with label choices sexist, non-sexist, NA \textit{depending on} explicitly provided values.




\subsection{Dataset}
We borrow multi-label sexism categorization data (multi-sexism) \cite{parikh-etal-2019-multi}, which offers fine-grained sexism categorization for sexist content.
Example categories include, but are not limited to, \textit{Role-stereotyping}, \textit{Pay gap}, and \textit{Mansplaining}. We select 10 items of content per category to have a small set of human-labeled data for the prompt-construction in our methodology and baselines. The rest of the data are used as the test set.

Based on the description of each category, we manually compose two opposing values -- one making the content  sexist (value) and another making the content  non-sexist (counter-value). For instance, any \textit{Role Stereotyping} contents will be considered to be sexist based on the value ``Men and women are equally capable for any role,'' but can also be considered to be non-sexist with the different value ``Men and women are biologically different; hence certain roles are more appropriate for women.'' A full list of values and counter values is available in Appendix \ref{appendix:value_c_value}. In total, we consider 19 categories of sexism and two corresponding values (value, counter-value) for each category, translated into 38 (19 $\times$ 2) human values.

\paragraph{Test set}
We use the original multi-label sexism content (human-labeled, non-synthetic) for creating a test set for the value-aligned judgement task, excluding that used for prompt-construction in the training data generation. Originally, each item of content is labelled with one/multiple sexism categories. For our task setup, we translate the data into the form of triplet \{content, value, label\}, and we assign value-dependent labels to each sample.
For instance, if content $C$ was originally labelled as \textit{Role-stereotyping} (\textit{RS}), we convert into three testing samples, \{$C$, value$_{RS}$, Sexist\}, \{$C$, counter-value$_{RS}$, Non-Sexist\}, and \{$C$, random value/counter-value, NA\}. Note that values for NA labels are totally unrelated to the content category. In this way, we can inspect the model's performance in making a value judgement on the same content with different values. In total, there are 17,720 test samples, with a label ratio of 1:1:1. 

\subsection{Models}
\subsubsection{\ours{s} (Ours)}

\paragraph{Generating value-aligned training data}
Using the method explained in Section~\ref{sec:data_generation}, we get 100 content pieces 
from each of the value and counter-value prompts. In sum, there are 200 unique pieces of content per category.\footnote{Reflecting the original ratio of multi-sexism, we keep the original number of samples if there are less than 100.} Then, all content per category is paired with a value and counter value and corresponding labels \{content, \textit{value}, `Sexist'\} and \{content, \textit{counter-value}, `Non-Sexist'\}. So, each content item has a duplicate but is paired with different values and value-aligned judgements. To prevent the model from only learning two value and label associations, we synthetically make the class `NA' by assigning irrelevant values/counter-values to the content (e.g., assigning the value of \textit{Pay Gap} to a content of \textit{Role Stereotyping} so the label is `NA'). In total, there are 10,722 samples, including the prompt construction samples. We split them into training and validation sets with a ratio of 4:1.

\paragraph{Building \ours{s}} We finetune smaller models with the generated value-aligned training data. We build \ours{s} to incorporate explicit human values to make judgements for value-aligned sexism classification following Section~\ref{sec:Fine-tuning_Classifiers}.
For the smaller models, we take base versions of ALBERT (12M params.) ~\cite{lan2019albert},  RoBERTa (125M params.)~\cite{liu2019roberta} and BART (110M params.)~\cite{lewis2019bart} to construct \vaalbert, \varoberta~and \vabart, respectively. 
RoBERTa has been proved to be robust in various NLP tasks and BART shows comparable performance to RoBERTa on GLUE tasks.



\subsubsection{Baselines}
\label{sec:Baselines}
To examine our proposed approach, we compare it with multiple baselines, including a random baseline, prompt-based few-shot learning with OPT-175B, and fine-tuning transformer-based models. For the fine-tuning setting, we fine-tune on different data setups -- only with human-labeled data (without generated data) and with semantically augmented data.

\paragraph{Random Baseline} We randomly select the predicted label for each test sample with the same label probability distribution as in the training data. 

\paragraph{OPT-175B (few-shot)} This baseline uses OPT-175B with a prompt-based few-shot learning for \textit{label prediction}.~\footnote{We use prompt-based few-shot learning with OPT-175B for generating \textit{value-aligned content} in our methodology while the baseline used it for directly predicting \textit{label}. Refer to Appendix \ref{sec:appendix_exp} for details.} We provide 20 few-shot samples in the context.

\paragraph{Human-Labeled (HL)-Models} We only use the small subset of human-labeled samples as training data to fine-tune smaller transformer-based LMs with a linear layer trained on top. We choose the base versions of ALBERT, RoBERTa and BART as the backbone models for a fair comparison with our \ours{s}.

\paragraph{Nlpaug-Models} Nlpaug~\cite{ma2019nlpaug} is semantic augmentation method using BERT-base embedding. We conduct augmentation with prompt construction examples by insertion and substitution. For each examples, we make 10 augmented samples (five insertions and five substitutions). Then, we fine-tune the base versions of ALBERT, BART and RoBERTa on the semantically augmented data and prompt-construction examples so we can evaluate the effectiveness of the prompt-based augmentation in our method.

\subsection{Experimental setup}
\label{sec:Experimental_setup}

\paragraph{Evaluation metric}
We evaluate our experiments with both F1 score and accuracy. For the main results, we report all accuracy, weighted F1-score (W-F1), precision and recall.

\paragraph{Implementation Details}
For generating value-aligned training data, we conduct the main experiment with OPT-175B model with top-p 0.7 and temperature 1. 
For our \ours{s} and \ftmodels~baselines, we use pre-trained transformer-based LMs available through the HuggingFace API.
Further implementation details such as hyperparameters are given in Appendix \ref{sec:appendix_exp}.

\begin{table}[t]
    \centering
    \begin{adjustbox}{width={0.45\textwidth},totalheight={\textheight},keepaspectratio}

    \begin{tabular}{l|cc}
    \toprule
    Model & $Accuracy$ & $W$-$F1$ \\ \midrule
    Random Baseline       & $33.53_{0.41}$ &  $33.53_{0.41}$\\ 
    OPT-175B (few-shot)   & $55.18_{7.75}$ & $54.78_{7.20}$\\
    \ftalbert  & $58.70_{4.43}$ & $51.67_{3.96}$ \\
    \ftroberta  & $64.53_{2.54}$ & $55.23_{1.91}$ \\
    \ftbart & $63.23_{1.87}$ & $54.93_{1.47}$ \\
    Nlpaug-ALBERT  & $62.87_{2.13}$ & $58.80_{3.44}$ \\
    Nlpaug-RoBERTa & $61.52_{3.03}$ & $58.67_{2.89}$ \\
    Nlpaug-BART & $59.03_{1.38}$& $58.49_{1.60}$ \\ \midrule
    \vaalbert & $70.10_{1.65}$ & $70.75_{1.48}$\\
    \varoberta & $73.24_{0.39}$& $73.82_{0.32}$\\
    \vabart & $\mathbf{74.07}_{0.82}$ & $\mathbf{74.36}_{0.60}$
    \\ \bottomrule
    \end{tabular}
    \end{adjustbox}
    \caption{Evaluation results of baselines and our proposed \ours{s}, on the value-alignment task. We use 200 value-aligned training data samples generated from the LLMs per category to fine-tune \ours. Experiments are ran with five random seeds and results are reported in $mean_{std}$ format. All our \ours~performances are statistically significant (t-test with $p$-value $<0.05$). Scores are all in percentage (\%).}
    \label{tab:main_results}
\end{table}

\section{Results and Analysis}
\label{results_and_analysis}


\subsection{Main Results}
\label{sec:main_results}
\paragraph{Effectiveness of our method}
Table \ref{tab:main_results} shows the performance of the models on the value-aligned sexism classification task. Our models achieve better scores on W-F1 and accuracy than the baselines by large gaps ($15.56\sim40.83$\% gain in W-F1), which signifies the robustness and superiority of our approach. Our \vaalbert~also surpasses all baselines, including those back-boned with bigger models (e.g., Nlp-RoBERTa, \ftroberta). This highlights the effectiveness of the value-aligned knowledge distillation with LLMs. 

We observe that the OPT-175B few-shot learning approach performs better than random label assignments on the test set and \ftalbert, but still performs worse than or as comparable as the other baselines. This indicates that LLMs with prompt-based few-shot learning can understand the value-aligned classification task to some extent, but the performance is still low. \ftmodels~surpass OPT-175B (few-shot) under all evaluation metrics except \ftalbert~in W-F1 score, showing that the models can capture our task with limited human-labeled data due to the effectiveness of fine-tuning. Nlpaug is one of the conventional data augmentation approaches and we augment the same amount of data as \ours. In comparison with \ftmodels, Nlpaug-models show higher W-F1 scores with small drops in accuracy.

Overall, the experimental results support our proposed approach for the value-aligned judgement task. OPT-175B (few-shot) shows much lower and unstable performance than \ours{s}~although the value-aligned training data of \ours{s}~is generated from OPT-175B. For the prompt-based few-shot approach, especially when the task setup is complicated like value-aligned classification, the model cannot easily overfit the task by giving several prompts, leading to a higher chance to predict random labels. Instead, we used a knowledge distillation approach through training data generation, which is a simpler task for the model as the main objective of the general language model is text generation.
Moreover, utilizing the LLMs for generating knowledge distilled data is more effective than simple semantic text augmentation (e.g., Nlpaug).

\begin{figure}[t]
    \centering
    \includegraphics[width=\linewidth]{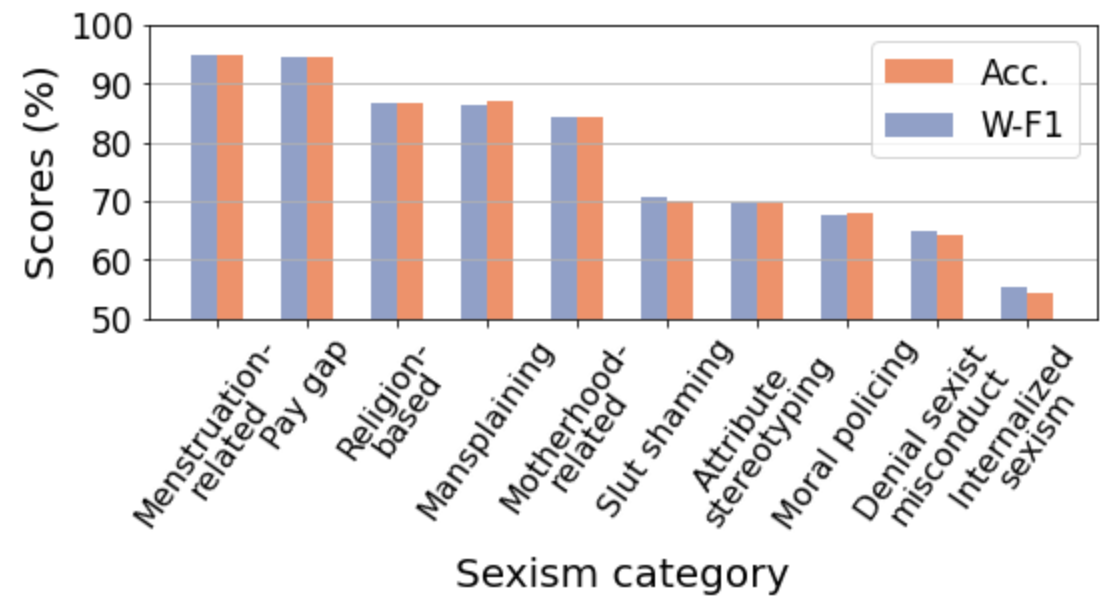}
    \caption{Evaluation results of \vabart~per sexism category on the test set. Only the top and bottom five categories (based on W-F1) are displayed. The performance for the nine categories in the middle are $\sim80\%$ for Acc. and W-F1. The full results for the 19 categories are available in Appendix~\ref{sec:appendix_Per_Category_Results}.}
    \label{fig:analyze_category}
\end{figure}

\paragraph{Per-Category performance} Figure~\ref{fig:analyze_category} presents the per-category evaluation scores of \vabart. The results vary significantly between categories, indicating the complexity of our proposed task. The results for both \textit{Menstruation-related Discrimination} and \textit{Pay Gap} achieve scores higher than 90\%, while the results for \textit{Internalized Sexism} are relatively low. We conjecture reasons for the high performance of certain categories are varying quality of generated training data per categories and more distinguishable features than other. We investigate this point further in Section~\ref{sec:quality_analysis}.

\subsection{Quality Analysis for Generated Value-Aligned Training Data}
\label{sec:quality_analysis}


\paragraph{Distinction between generated data \& test set}
The vocabulary overlap between all generated data (training set for \ours{s}) and test set data is 51.79\%.
Moreover, we check how many of generated data samples that share more than 80\% of vocabulary with at least one of the test data samples, finding that \textit{only} 0.01\% of generated data samples reach the threshold (80\%). Therefore, the data generated from OPT-175B for training \ours{s}~is clearly distinct from the test set.

\begin{figure}[t]
    \centering
    \includegraphics[width=0.8\linewidth]{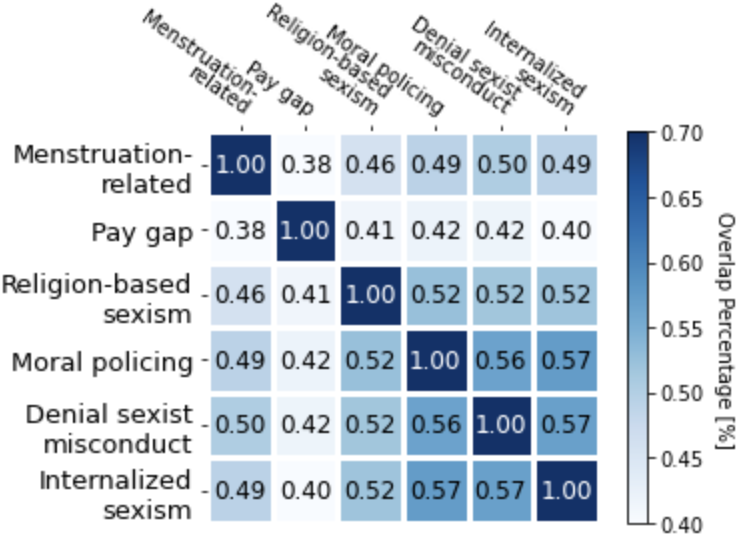}
    \caption{Vocabulary overlaps (\%) of the generated data among sexism categories. Only top-3 and bottom-3 categories are displayed in descending order of W-F1 (top to bottom; left-to-right). Full set is in Appendix~\ref{sec:appendix_vocabulary_overlaps}.}
\label{fig:analyze_overlap}
\end{figure}

\paragraph{Diversity of Data}
We calculate the vocabulary overlaps for each sexism category of the generated data in Figure~\ref{fig:analyze_overlap}. We observe that the vocabulary overlaps are generally small, which illustrates that OPT-175B can generate diverse data for different values (e.g., sexism categories) provided in prompts. We can observe the trend that the overlaps among high performing categories are small, especially \textit{Pay gap} and \textit{Menstruation related}, which make data sample distinguishable to others. In contrast, low performing categories, overlaps are relatively higher.


\paragraph{Human evaluation}
LLMs are powerful few-shot learners, yet they are not perfect. Thus, we conduct human evaluation on two categories' data (\textit{Internalized sexism} and \textit{Pay gap}) to further investigate the augmented data quality. We assess generated contents from two aspects: 1) relevance to the corresponding category ($R$); 2) sexism ($S$).
Pay gap data are evaluated to be relevant and sexist ($R$\&$S$) 69.44\% of the time ($R$: 75.93\%, $S$: 87.03\%) while internalized sexism data are evaluated to be 25\% ($R$\&$S$) ($R$: 34.50\%, $S$: 67.50\%).
We observe that the quality of \textit{Pay gap} generated data is much better than that of \textit{Internalized sexism}, which is consistent with the per category results in Figure~\ref{fig:analyze_category}. This highlights the difficulty of our task and the need for more robust prompt templates for prompt-based data generation. And the human-in-the-loop method may further boost the performance of our approach with less noisy data.

\paragraph{Effectiveness of generated training data}
\begin{table}[t]
    \centering
    \begin{adjustbox}{width=0.85\linewidth,totalheight={\textheight},keepaspectratio}
    \begin{tabular}{l|cc}
    \toprule
    Model & $Accuracy$ & $W$-$F1$  \\ \midrule
    \vaalbert  & $70.10_{1.65}$ & $70.75_{1.48}$ \\
        \qquad w/o human labeled data & $70.79_{1.40}$ & $71.29_{1.33}$\\\midrule
    \varoberta & $73.24_{0.39}$& $73.82_{0.32}$\\
        \qquad w/o human labeled data & $72.90_{2.06}$ & $73.19_{1.68}$ \\\midrule
    \vabart            & $\mathbf{74.07}_{0.82}$ & $\mathbf{74.36}_{0.60}$\\
        \qquad w/o human labeled data &  $72.30_{1.24}$& $72.71_{0.90}$ \\
    \bottomrule
    \end{tabular}
    \end{adjustbox}
    \caption{Effectiveness of generated data. We remove human-labelled data from the training set and \textbf{only} use synthetic samples generated from LLM for training (w/o human-labeled data). The minimal drops in performance show the effectiveness of value-aligned training data generated from LLMs for the value alignment task.}
    \label{tab:ablation_study}
\end{table}

To investigate the standalone effectiveness of the generated training data (value-aligned knowledge distillation), we study the performance of \ours{s}~when they are trained \textit{without} any of human-labeled data but \textbf{only} with generated data (Table \ref{tab:ablation_study}). Minor performance degradations in both \vabart~and \varoberta~are investigated, $-1.65\%$ and $-0.63\%$ W-F1 respectively. However, these values are still above those of the baselines. Interestingly, \vaalbert~showed a minimal performance gain on both accuracy and W-F1. This indicates that the value alignment knowledge distilled from LLMs is the main contributor for \ours~to understand the task.


\begin{table}[t]
    \centering
    \begin{adjustbox}{width=0.8\linewidth,totalheight={\textheight},keepaspectratio}
    \begin{tabular}{l|cc}
    \toprule
    Model & $Accuracy$ & $W$-$F1$ \\ \midrule
    OPT-175B (fewshot) & $32.97$ & $30.23$ \\
    \ftalbert  &  $40.25_{6.05}$& $37.52_{7.68}$ \\
    \ftroberta  & $47.79_{5.65}$& $45.35_{6.09}$ \\
    \ftbart &  $46.09_{3.42}$& $45.97_{3.68}$ \\
    
    Nlpaug-ALBERT  &  $48.10_{11.0}$& $40.62_{8.05}$\\
    Nlpaug-RoBERTa &  $40.14_{2.00}$& $30.64_{3.56}$\\
    Nlpaug-BART &  $47.76_{3.89}$& $42.40_{5.45}$ \\ \midrule
    
    \vaalbert &  $55.15_{6.83}$& $53.14_{9.05}$ \\
    \varoberta &$\mathbf{58.13}_{5.33}$& $\mathbf{56.60}_{6.56}$\\
    \vabart & $57.98_{5.12}$& $55.94_{5.48}$
    \\ \bottomrule
    \end{tabular}
    \end{adjustbox}
    \caption{Performances of \ours{s}~and baselines on \textbf{unseen} values in value-aligned sexism classification. In the training phase, models did not see any of the values in the test set.}
    \label{tab:unseen}
\end{table}

\subsection{Generalization Ability on Unseen Values}

To understand capacity of models to generalize value-aligned judgement over unseen values, we conduct an experiment in which three randomly selected sexism categories are separated from the training process (i.e., models have never seen values related to the three categories in the training phase and are evaluated on test set only composed of those unseen values) and the results are presented in Table~\ref{tab:unseen}. 
Overall, there are drops in performance compared to the main experiment (Table~\ref{tab:main_results}), while all of our \ours{s} continue to outperform all baselines. The baselines experience larger drops (maximum 43.18\% for Nlpaug-RoBERTa) than the \ours{s} (17.22\% for \varoberta). Considering the model was never taught or received any direct supervision on the test values, it is expected behavior as other generalization problem. We leave how to improve the models' generalization ability in value-aligned judgement task for future work.




\subsection{Ablation Studies}
\label{sec:ablation_study}

\paragraph{LLMs capacity for prompting}
\label{sec:LLMs'_capacity}
We first investigate how the size of LLMs affects the capacity for generating value-aligned training data by evaluating the final performance of \ours~trained on data from varying sizes of LLMs.
Unsurprisingly, as is shown in Table \ref{tab:size_of_llms}, we can continually boost the model's performance when the LLMs size increase.

We also train \vabart~with the data prompted from GPT-Jurassic. Results for GPT-Jurassic 6B are slightly higher than those of OPT-6.7B, although the model size is smaller. However, when the LLMs become extremely large, GPT-Jurassic 178B performs similar to OPT-175B with only 0.12\% difference. Since similar model sizes show similar performance with minimal differences, the types of LLMs do not have much effect on the generated data quality for our task.



\begin{table}[t]
    \centering
    \begin{adjustbox}{width=0.95\linewidth,totalheight={\textheight},keepaspectratio}
    \begin{tabular}{l|cc}
    \toprule
    Models & $Accuracy$ & $W$-$F1$ \\ \midrule
    \vabart (OPT-1.3B) & $65.72_{2.03}$ & $66.50_{2.15}$ \\
    \vabart (OPT-6.7B) &  $65.69_{2.28}$ & $66.44_{2.45}$ \\
    \vabart (OPT-175B) & $\mathbf{74.07}_{0.82}$ & $\mathbf{74.36}_{0.60}$\\ \midrule
    \vabart (GPT-Jurassic 6B) & $69.09_{1.59}$ & $69.89_{1.49}$ \\
    \vabart (GPT-Jurassic 17B) & $71.03_{1.01}$ & $71.68_{0.90}$ \\
    \vabart (GPT-Jurassic 178B) & $74.04_{1.16}$ & $74.24_{0.91}$ \\\bottomrule
    \end{tabular}
    \end{adjustbox}
    \caption{Effect of size and types of LLMs 
    on value-aligned training data generation. 
    We prompted OPT and GPT-Jurassic ranging 1.3B $\sim$ 178B. The bigger the model, the better the final performance in the value alignment task. All \vabart~variations are fine-tuned with the same number of training samples.}
    \label{tab:size_of_llms}
\end{table}

\begin{figure}[t]
    \centering
    \includegraphics[width=\linewidth]{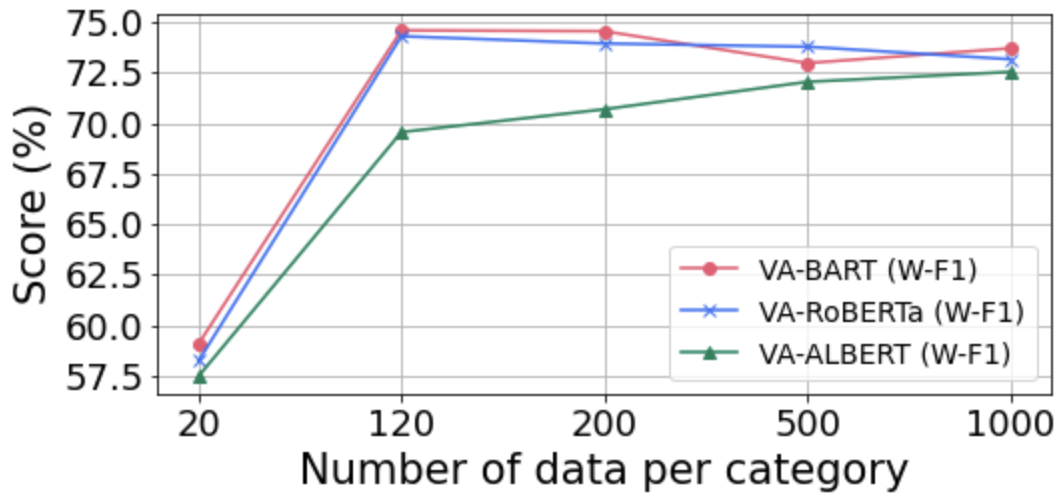}
    \caption{Evaluation results (W-F1) of \ours{s}~over different size of generated training data.}
    \label{fig:data_size}
\end{figure}

\paragraph{Effect of the size of generated training data}
\label{sec:number_of_data}
To investigate whether increasing the the number of generated data can gain further improvements, we fine-tune \ours{s}~with different training data size. 
In Figure \ref{fig:data_size}, we show that the W-F1 score does not show any gain when the size exceeds 200 samples except for \vaalbert. As we analysed in Section~\ref{sec:quality_analysis}, the generated data has noise. We conjecture that when using more generated data, the additional data will not only bring more value alignment knowledge, but also add more noise to the training set. Therefore, when the degradation in model performance caused by the noisy data is greater than the improvement in model performance from the additional knowledge, the overall results decrease.

\section{Conclusion and Future Work}
\label{sec:Conclusion_and_Future_Work}
In this paper, we propose a task that focuses on teaching a model human value alignment knowledge. We also introduce value-aligned models (\ours) that generate value-aligned training data from LLMs by prompt-based data generation and fine-tune smaller classification models with the value-aligned generated training data. Experimental results show that \ours~generally outperforms strong baselines.
Further analysis illustrates that the generated data from larger LLMs helps increase the performance, and more generated data can cause performance reduction when the data size is too large. In addition, we also test the promising generalization ability of \ours. Finally, we highlights several research challenges for future work: improvements in 1) the robustness of the model on diverse values, 2) the models’ generalization ability for our value-aligned judgement task, 3) higher quality generated data with more human curation.

\section*{Limitations}



Our methodology is currently tested with only English. We conjecture that the methodology should be applicable to other languages, but may be limited by the capacity of LLMs in those specific languages. It is possible that value-aligned knowledge distillation may be more difficult with languages from countries and regions that do not have a complete set of human value definitions. Thus, exploring the value-aligned task in different languages other than English is a promising research direction. 

Our main experimental results are based on a 175B parameter model, which requires large GPU resources or access through an API. This may hinder other researchers from reproducing experimental results. Additionally, we explored different sizes of LLM including 1B and 6B models, which do not require large GPU resources, and showed they can achieve comparable results. We hope they can be possible alternative options for researchers who may not have access to 100B+ models.


Although sexism is a suitable case study for us to investigate the feasibility of the value alignment task as we have shown throughout this work, it is still one domain. Further expansion to different domains or value-aligned classification tasks such as the detection of racism, toxicity, other than sexism, are needed.

\section*{Ethics Statement}
In this work, we conduct experiments with some values that are unconventional in and counter to the current contemporary society. However, we are not suggesting tolerance on sexist behaviors or beliefs. Instead, we explain the existence of different perspectives in the discussion of sexism across cultures or religions. Our value-aligned sexism classification task is a case study of this decoupled process. 

We propose that the human value definition should be decoupled from the value-alignment task. The former should be defined together with society, ethicists, social scientists, and so on. However, the mechanics of value alignment should be independent of the first part so that engineers do not define these values directly in the training data or in the code itself.





\bibliography{custom}
\bibliographystyle{acl_natbib}
\newpage
\appendix
\section{Additional Data details}
\subsection{Value and Counter Value}
\label{appendix:value_c_value}
In total, we have 19 categories of sexism and two corresponding values for each of them. We provide the list of all values and counter-values per sexism category in Table~\ref{tab:values_list}.

\section{Prompt Template Choice for Value-aligned Data Generation}
\label{appendix:Prompt_Choice_and_Results}
\paragraph{Final Prompt}
For prompting the value-aligned training samples, we selected the final prompt template out of a set-of five prompt templates based on experimental results with the OPT-175B model. The prompt template consists of an instruction header and an enumeration of content as follows:
\texttt{Generate \{label\} content that is relevant to the Value. Value:\{value\}$\backslash$n}. 

Here the label and value in brackets is replaced with the target label, and we provide five content examples with format \texttt{Content:\{content\}}, concatenated to the previous prompt. For these examples, We manually create the ten most representative examples that align with the described value and randomly select five of them for each prompt. Then, the model is encouraged to generate content relevant to the provided value and label it with the prompt \texttt{Content:}.

\paragraph{Tried Prompt Templates}
We tried five prompt templates, including the final prompt template as follows:
\begin{enumerate}
    \item \texttt{Generate \{label\} content that is relevant to the Value. Value:\{value\}$\backslash$n}.
    \item \texttt{``Each item in the following list contains a value and the respective "\{label\}" content according to the value.Value:\{value\} Content:\{content\}''}
    \item \texttt{``value="\{value\}"$\backslash$n label="\{label\}"$\backslash$n content=\{content\}''}
    \item \texttt{``Value:\{value\} Label:\{label\} Content:''}
    \item \texttt{``Generate {label} content that is relevant to the Value.$\backslash$nValue:\{value\} Content:\{content\}''}
\end{enumerate}

We mainly investigated the effectiveness of the different prompt templates with the OPT-175B model as we conducted the main experiment with it. We also did investigation with the GPT-Jurassic 6B model. Interestingly, the GPT-Jurassic models showed better performance with data prompted with prompt template \#2, which was different from OPT-175B. This may have resulted from the different training objectives and pre-training resources of the models. Although the overall structure of our methodology is model agnostic, there should be some exploration made on prompt template construction dependent on models.

The experimental results are shown in Table \ref{tab:promptype}.

\begin{table}[h]
    \centering
    \begin{adjustbox}{width={0.48\textwidth},totalheight={\textheight},keepaspectratio}
    \begin{tabular}{rccccc}
    \toprule
\multicolumn{1}{l}{\begin{tabular}[c]{@{}l@{}}\varoberta~\\w/ prompt type\end{tabular}} & $Acc.$ & $Prec.$ & $Rec.$ & $W$-$F1$  \\ \midrule
1 & \textbf{73.91\%} & 73.91\% & 75.14\% & \textbf{74.31\%} \\
2 & 72.71\% & 72.71\% & 75.22\% & 73.34\% \\
3 & 71.25\% & 71.25\% & 75.48\% & 72.06\% \\
4 & 69.75\% & 69.75\% & 74.12\% & 70.60\% \\
5 & 72.07\% & 72.07\% & 73.82\% & 72.61\% \\ \bottomrule
\end{tabular}
    \end{adjustbox}
        \caption{Evaluation results of \varoberta~trained on OPT-175B generated 
data with different prompt types. We prompted 120 data samples per categories.}
    \label{tab:promptype}
    \end{table}

\begin{table*}[]
    \centering
    \begin{adjustbox}{width=\linewidth,totalheight={0.9\textheight},keepaspectratio}
    \begin{tabular}{p{0.22\linewidth}p{0.5\linewidth}p{0.5\linewidth}}
    \toprule
    Category & Value & Counter-value \\ \midrule
    Role stereotyping & Men and Women are equally capable for any role. & Men and women are biologically different hence certain roles are more appropriate for women. \\\midrule
    Attribute stereotyping & The differences between people's physical, psychological, or behavioral qualities or likes/dislikes are not based on gender. & Women and men like/dislike different things and they behave differently because they have been raised and educated differently. \\\midrule
    Body shaming & Objectionable comments or behaviour concerning appearance including the promotion of certain body types or standards are not acceptable. & Women and men have different body types or standards and they should be promoted. \\\midrule
    Hyper-sexualization (excluding body shaming) & Unwarranted focus on physical aspects or sexual acts to women is not acceptable. & Attention and appreciation of the female physical aspects or sexual acts is acceptable. \\\midrule
    Internalized sexism & Women making sexist comments and actions are not acceptable. & If women make these comments or actions then they are not sexist.; Women cannot be sexist. \\\midrule
    Pay gap & Salaries for men and women for the same work profile should be equal. & Women are paid less than men because they contribute less than men at work. \\\midrule
    Denial or trivialization of sexist misconduct & None of sexist comment or behavior is acceptable. & Some comments or behaviors can be considered banter. \\\midrule
    Threats & Anything that makes women feel threatened is inappropriate. & Some threats to women can be seen as a joke. \\\midrule
    Sexual assault\quad(excluding rape) & Any sexual contact should happen with explicit consent. & Willingenss for sexual contact is implied by things like past behavior, what you wear, or where you go. \\\midrule
    Sexual harassment\quad\quad(excluding assault) & Any sexually objectionable behavior to women is not acceptable. & Some culturally and traditionally based actions against women cannot be considered sexual harassment. \\\midrule
    Tone policing & Comments or actions that cause or aggravate restrictions on how women communicate is not allowed. & Women should pay attention to their language in communication. \\\midrule
    Moral policing (excluding tone policing) & Moral principles should be the same for Women and Men. & There are different codes of conduct for women and men in different cultures and religions. \\\midrule
    Victim blaming & Victims does not hold responsibility for any of sexual harassment, violence, or other sexism perpetrated against her. & In some cases, victims are responsible for sexual harassment, violence, or other sexism perpetrated against her. \\\midrule
    Slut shaming & Individual sexual freedom should be respected and accepted.; Women are free to wear whatever they want. & Sex should be more subject to self-control.; Women should be more conservative in terms of sexual conduct and behavior.; Women should dress in a way that is not linked to sexual availability. \\\midrule
    Motherhood-related discrimination & Shaming, prejudices, or other discrimination or misconduct related to the notion of motherhood is not acceptable. & Motherhood is honourable, women should be proud of this. A woman should be a mother. \\\midrule
    Menstruation-related discrimination & Shaming, prejudices, or other discrimination or wrongdoings related to menstruation is not acceptable. & Women should understand that they are often more emotional during menstruation due to hormones.; Women should not enter certain venues during menstruation.; Women should be exempted from difficult tasks when they are in menstruation. \\\midrule
    Religion-based sexism & Sexist discrimination or prejudices stemming from religious scriptures or constructs is not allowed. & Women should respect and follow religious scriptures or constructs. \\\midrule
    Physical violence (excluding sexual violence) & Physical violence against anyone is objectoinable. & Women need to be physically punished for certain misbehavior, such as adultery. \\\midrule
    Mansplaining & Giving unsolicited advice or explanation to women for something that they actually know well or disapprove is not acceptable. & Women should humbly consider the advice of others, whether it is appropriate or not. \\ \bottomrule
    \end{tabular}
    \end{adjustbox}
    \caption{List of the values and counter-values for each sexism category.}
    \label{tab:values_list}
\end{table*}

\section{Experimental Details}

\label{sec:appendix_exp}
\paragraph{Hyperparameters}
For hyper-parameters, we perform a grid search to find the best performing set of parameters among the learning rates \{1e-5, 5e-5\} and batch sizes \{32, 64\}. 

\paragraph{Training Details}
For each model we train for a maximum of 30 epochs with early-stopping with patience of 5. Each experiment is conducted on an Nvidia RTX 3090 device, and each epoch takes around 2--10 minutes depending on the number of training samples. 

\paragraph{Random Seeds}
We ran each of experiments five times with different random seeds and reported the mean and standard deviation in a format of $mean_{std}$. 

\paragraph{OPT-175B (few-shot) Baseline Prompt} For each test sample, we construct a prompt with the task instruction and several examples as shown below:

\texttt{``Predict a Label for the Content based on the given Value: \textbf{Value}. Content: \textbf{Content} Label: \textbf{Label} $\backslash$n Predict a Label for the Content based on the given Value: \textbf{Value}. Content: \textbf{Content} Label:''}

In the prompt, the bold words will be replaced by the actual data. The first sentence is the few-shot example and we repeat it $N$ times by randomly selecting five samples for each label category. The second sentence is the test sample, and the model will generate the corresponding label in the text. During generation, we set top-p 0.9 and generate labels five times. Finally, we calculate the average scores among the results.

%



\section{Vocabulary overlaps of generated training data among sexism categories}
\label{sec:appendix_vocabulary_overlaps}

\begin{figure*}[t]
    \centering
    \includegraphics[width=\linewidth]{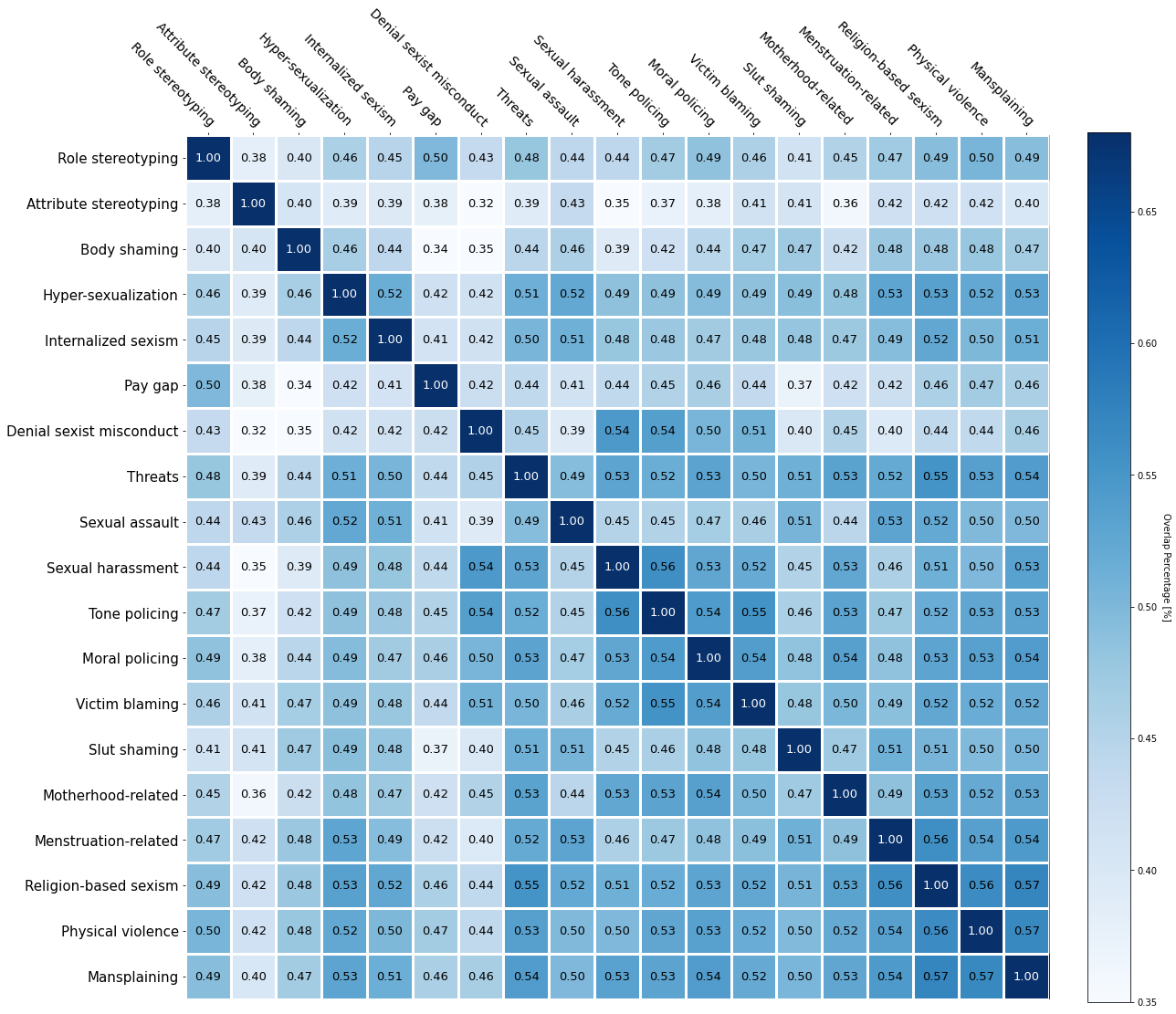}
    \caption{Vocabulary overlaps of the value-aligned training data generated from OPT-175 among sexism categories. Categories are in descending order of W-F1 (top to bottom; left-to-right).}
    \label{fig:analyze_overlap_all}
\end{figure*}

Figure~\ref{fig:analyze_overlap_all} presents the vocabulary overlaps of the value-aligned training data generated from OPT-175 among the sexism categories. We calculate the vocabulary overlaps for each sexism category of the generated data.



\section{Per Category Results}
\label{sec:appendix_Per_Category_Results}
\begin{figure*}[]
    \centering
    \includegraphics[width=\linewidth]{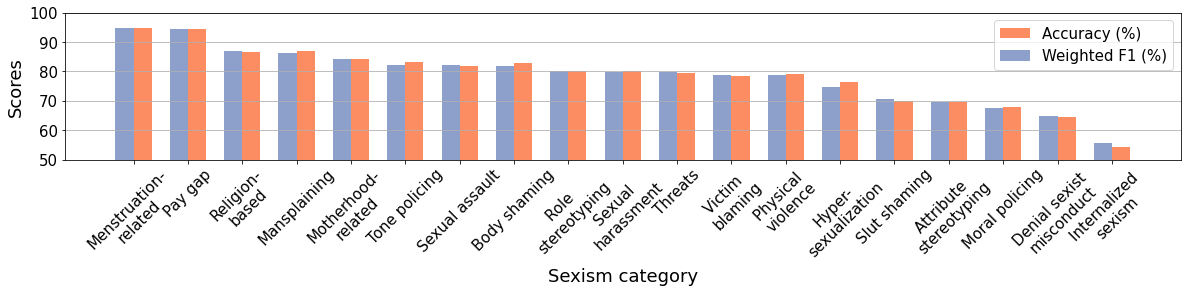}
    \caption{Evaluation results of \vabart for each sexism category on the test set.}
    \label{fig:analyze_category_full}
\end{figure*}

Figure~\ref{fig:analyze_category_full} presents the evaluation results of \vabart~for each sexism category on the test set.



\end{document}